\documentclass[10pt,twocolumn,letterpaper]{article}

\usepackage[pagenumbers]{cvpr}
\definecolor{cvprblue}{rgb}{0.21,0.49,0.74}
\usepackage[pagebackref,breaklinks,colorlinks,allcolors=cvprblue]{hyperref}
\usepackage[T1]{fontenc}
\usepackage{graphicx,verbatim}
\usepackage{amssymb}
\usepackage{amsmath}
\usepackage{graphicx}
\usepackage{textcomp}
\usepackage{amsmath}
\usepackage{stfloats}
\usepackage{url}
\usepackage{verbatim}
\usepackage{multirow}
\usepackage{booktabs}
\usepackage{color}
\usepackage{xcolor}
\usepackage{algorithm}
\usepackage{algorithmic}
\usepackage{caption}
\usepackage{hyperref}
\usepackage{makecell}
\usepackage{multirow}
\usepackage[normalem]{ulem}
\usepackage{pifont}
\usepackage{placeins}
\usepackage{array}
\usepackage{multicol}
\usepackage{colortbl}
\usepackage{threeparttable}
\usepackage{array}
\def\BibTeX{{\rm B\kern-.05em{\sc i\kern-.025em b}\kern-.08em
    T\kern-.1667em\lower.7ex\hbox{E}\kern-.125emX}}
\def\confName{CVPR}
\def\confYear{2027}

\title{One-shot is Enough: Boosting SAM for Medical Image Segmentation through Structured Information Modeling}
\author{
Guilin Guan$^{*}$\\
Dalian University of Technology\\
{\tt\small guanguilin@mail.dlut.edu.cn}
}

\begin{document}

\maketitle
\begingroup
\renewcommand{\thefootnote}{*}
\footnotetext{Corresponding author.}
\endgroup
\begin{abstract}
In recent years, Foundation models such as the Segment Anything Model (SAM) have demonstrated strong generalization capabilities, but because they are primarily pre-trained on natural images, they perform poorly when faced with medical images. With sufficient labeled medical data, this challenge is somewhat alleviated. However, the difficulty and high cost of acquiring large amounts of labeled data make one-shot learning a highly attractive research paradigm. But within this paradigm, SAM struggles to fully extract its potential multi-granular information, leading to a sharp decline in performance. To address this, we propose OB-SAM, a one-shot enhancement framework for medical image segmentation. It aims to maximize the utilization of information from a single sample from the perspective of structured modeling. Specifically, we first designed a Multi-Frequency Decoupling (MFD) module, which divides features into different frequency domain components, allowing each branch to focus on information expression at different semantic levels, thereby alleviating overfitting and enhancing feature representation capabilities. Secondly, we introduce a Progressive Feature Integrator (PFI) module to promote complementary fusion among multi-level features, thereby avoiding the dominant effect of a certain level of feature in the fusion process. Furthermore, we propose an Selective Context Filtering (SCF) Module that adaptively selects key contextual information by modeling the contrast relationship between the foreground and background, thereby achieving precise localization of the lesion area. Extensive experiments across seven diverse datasets validate that OB-SAM achieves state-of-the-art performance.
\end{abstract}    
\section{Introduction}
\label{sec:intro}

\begin{figure}[h]
\centerline{\includegraphics[width=\columnwidth]{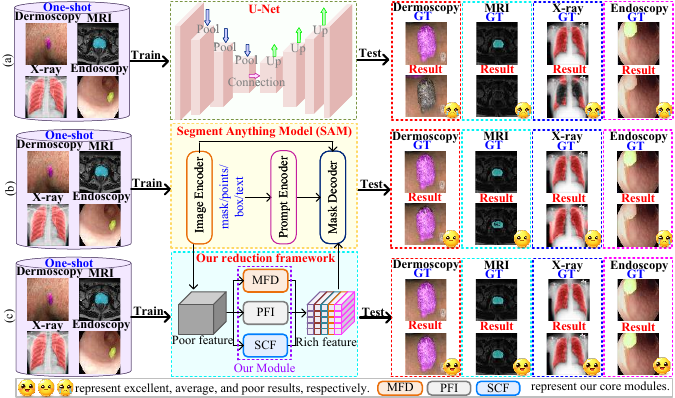}}
\caption{The segmentation results of U-Net, SAM, and OB-SAM in four modalities were used. These results demonstrate that OB-SAM has excellent generalization ability on medical data.}
\label{fig1}
\end{figure}
Medical image segmentation (MIS) is a core task in medical image analysis, aiming to accurately extract regions of interest (ROIs) from complex and highly heterogeneous backgrounds \cite{shaker2024unetr++,rahman2024emcad,ruan2024vm,wang2022medical}. High-precision segmentation results are crucial for clinical applications, as even subtle segmentation errors can significantly impact subsequent decision-making. The emergence of deep learning methods such as convolutional neural networks (CNN) has significantly accelerated progress in this field, thanks to large-scale, finely labeled datasets \cite{ronneberger2015u,zhou2018unet++,isensee2021nnu}. However, obtaining such labeled datasets remains a major bottleneck due to their high cost, time-consuming nature, and reliance on expert annotation \cite{bateson2022source}. Against this backdrop, one-shot MIS has gained increasing attention, offering a promising solution to alleviate the data dependency problem due to its learning capabilities with minimal labeled samples.\\
\indent Despite its significant potential in mitigating reliance on data annotation, one-shot MIS still faces substantial challenges in practical applications. First, under extreme conditions with only a single labeled sample, the model struggles to adequately depict the complex and varied anatomical structures and lesion morphologies in medical images, resulting in a marked deficiency in modeling intra-class differences and inter-class similarities \cite{mann2022endoscopic}. Second, the high heterogeneity among different patients in imaging modalities, tissue structures, and lesion manifestations makes the model prone to overfitting the training samples, hindering effective generalization \cite{karthik2022contour}. Furthermore, lesion regions in medical images are often small in scale and have low contrast with surrounding tissues, further exacerbating the difficulty of feature learning \cite{bhandary2023investigation}. Traditional deep convolutional methods rely on large amounts of data to learn robust feature representations; under one-shot settings, they struggle to extract discriminative lesion semantic information from limited supervised signals, leading to significant degradation in segmentation performance. As shown in Fig. \ref{fig1}(a), convolutional neural networks are almost ineffective at identifying lesion regions in one-shot scenarios, which significantly limits the potential of one-shot MIS in actual clinical applications.\\
\indent In recent years, large-scale foundational models, represented by SAM, have marked a paradigm shift in the field of computer vision, with their zero-shot generalization capabilities bringing new possibilities to downstream tasks \cite{kirillov2023segment}. However, SAM is pre-trained on natural images, which have significant domain differences from medical images, and its direct application to MIS still faces severe challenges. To achieve reliable segmentation performance, fine-tuning with large-scale medical data is usually required, but due to the difficulty in obtaining medical data, efficient adaptation based on one-shot settings is more practical. Although SAM has powerful feature representation capabilities, it is difficult to fully mine the multi-granular semantic information in medical images under conditions of very few labeled samples, resulting in a significant performance drop, as shown in Fig. \ref{fig1}(b). This limitation is mainly reflected in the following three aspects: First, under single-sample constraints, the model is prone to overfitting and it is difficult to decouple structural information at different scales, making it difficult to balance global and local features; second, SAM lacks a fine-grained balancing mechanism in multi-level feature fusion, and features at one level may dominate prediction, suppressing complementary information at other levels; finally, SAM is difficult to adaptively select key contextual information from limited annotations, and the contrast relationship between foreground and background cannot be fully modeled, making the segmentation results susceptible to background interference.\\
\indent To address the limitations of general-purpose visual models in one-shot applications, we propose the OB-SAM framework. This framework introduces multi-frequency decoupling, progressive feature Integrator, and selective context filtering strategies to maximize the information utilization of a single sample from a structured modeling perspective and exhibiting powerful segmentation performance, as shown in Fig. \ref{fig1}(c). Our main contributions are as follows:
\begin{itemize}
    \item We propose OB-SAM, a general medical image segmentation framework based on pre-trained SAM, which can be fine-tuned with only a single sample and has robust generalization capabilities.
    \item We introduced three innovative modules from the perspective of structured modeling: (1) Multi-Frequency Decoupling: By transforming features into three frequency domains: Fourier, wavelet, and Gabor, the Fourier domain focuses on global structure and periodic pattern modeling, the wavelet domain focuses on multi-scale local features and texture mining, and the Gabor domain focuses on multi-directional orientation and detail sensitivity enhancement. The three domains work together to achieve semantic-level information decoupling, alleviate the risk of overfitting in single-sample scenarios, and enhance feature expression capabilities; (2) Progressive Feature Integrator: By constructing a bidirectional interaction mechanism between multi-level features, progressive complementary fusion of features is promoted, effectively suppressing the dominant effect of single-level features in the fusion process, ensuring the balance of feature expression and the full utilization of information; (3) Selective Context Filtering: By explicitly separating the input feature map into foreground and background features, and optimizing features through independent enhancement branches, the network can adaptively filter key context information, suppress redundant background interference, and ultimately achieve accurate localization of lesion areas. 
    \item Numerous experiments have shown that OB-SAM consistently outperforms state-of-the-art baselines on seven datasets, achieving powerful generalization capabilities using only a single training data set, highlighting its practicality in data-scarce scenarios.
\end{itemize}

\section{Related Work}
\label{sec:formatting}
\subsection{Medical Image Segmentation}
Medical image segmentation is a fundamental task in image-guided diagnosis and treatment. Convolutional Neural Networks (CNN) have laid the baseline framework for this field \cite{ronneberger2015u}, and subsequent studies have continuously improved their segmentation accuracy and structural adaptability through strategies such as residual connections \cite{alom2018recurrent,mubashar2022r2u++}, attention mechanisms \cite{cai2022ma,zhang2025advances}, loss function improvements \cite{thanh2021crf,nasalwai2021addressing}, and generative adversarial training \cite{mondal2018few,nie2020adversarial}. However, CNN methods generally rely on large-scale, high-quality labeled data for training and are limited by local receptive fields, easily producing semantically inconsistent fragmented outputs, which severely restricts their generalization ability and clinical application scope in scenarios with scarce annotations.
\subsection{Few-shot Segmentation}
Few-shot segmentation (FSS) has attracted much attention because it can segment anatomical structures with only a small number of labeled samples. To this end, recent studies have proposed a variety of strategies, including context modeling \cite{tang2021recurrent}, global relevance enhancement \cite{sun2022few}, spatial prior guidance \cite{yu2021location}, query-guided meta-segmentation \cite{shen2023q}, and anatomical knowledge embedding networks \cite{wei2021akfnet}, which have significantly improved the accuracy of few-shot medical segmentation. Despite these advances, the generalization ability of these methods is still severely limited by specific training tasks or datasets, and their performance drops significantly when faced with unseen target categories.
\subsection{SAM-based Segmentation}
The SAM \cite{kirillov2023segment} has demonstrated superior zero-shot performance on natural images, thanks to its large training dataset and prompt-driven architecture. Inspired by its generalization potential, researchers have explored how to fine-tune SAM for medical image segmentation. Currently, strategies such as multi-box prompts \cite{deng2023sam}, uncertainty-guided adaptation \cite{xie2024masksam}, hybrid CNN-ViT prompt decoder \cite{gowda2024cc}, low-rank adaptation \cite{mahla2025medsaga}, category-aware prototype generation \cite{yan2025pgp}, and bidirectional attention mechanisms \cite{zhou2024biasam} have been proposed to improve its performance on medical tasks. However, despite these advances, applying SAM to one-shot scenarios remains challenging, primarily because the model struggles to fully extract the multi-granular information contained in medical images under this setting.
\section{Method}
\begin{figure*}[!t]
\centerline{\includegraphics[width=2.05\columnwidth]{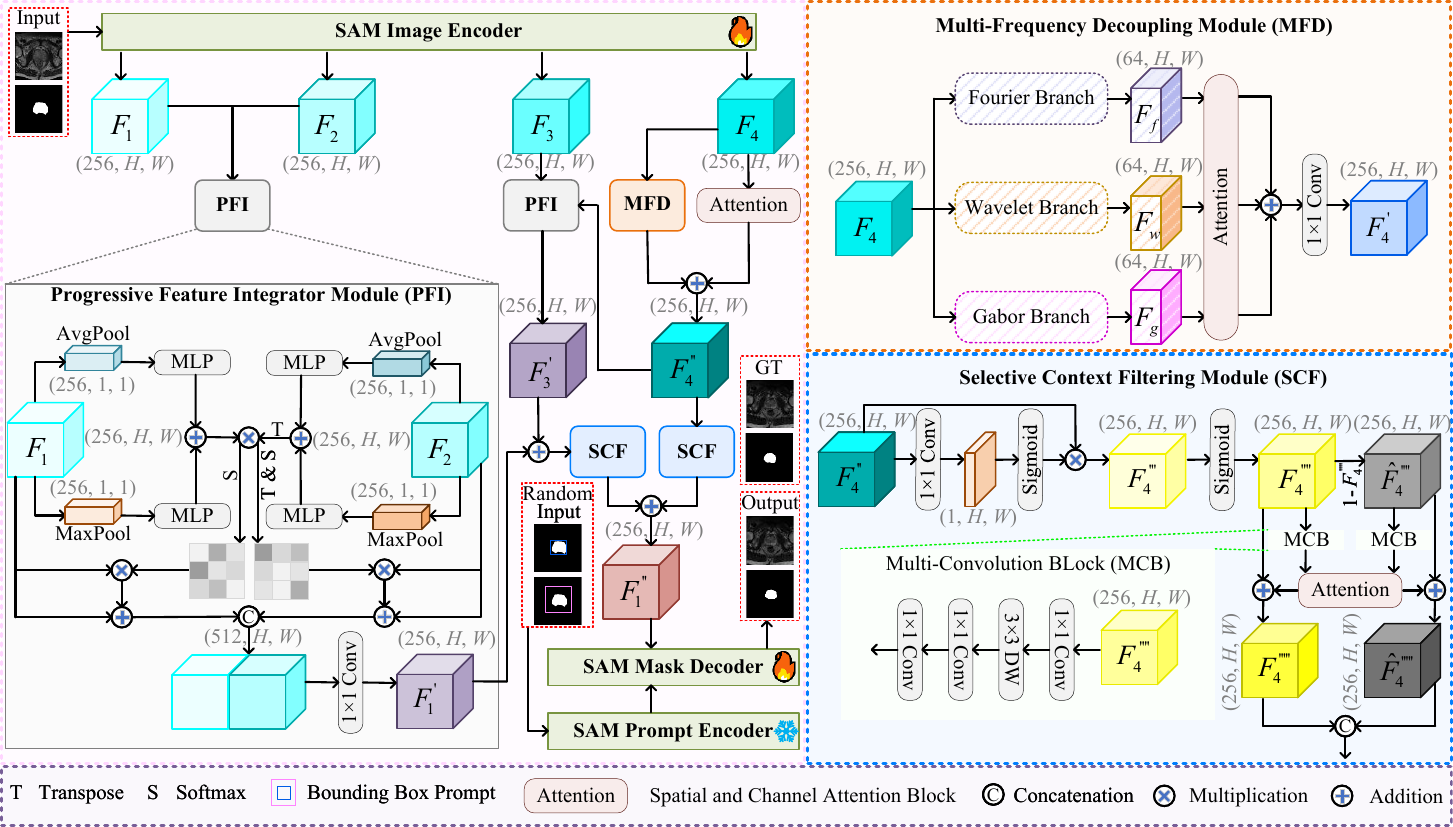}}
\caption{Overview of the proposed OB-SAM framework. The SAM image encoder extracts multi-level features. MFD captures complementary frequency information, PFI promotes cross-level feature fusion, and SCF enhances foreground representations while suppressing background interference. The refined features are fed into the SAM mask decoder for segmentation.}
\label{fig2}
\end{figure*}
This section elaborates on our proposed OB-SAM, a one-shot enhancement framework for medical image segmentation based on the SAM architecture. The OB-SAM architecture revolves around three key components: (1) MFD, used to alleviate overfitting and enhance feature representation capabilities; (2) PFI, used to promote complementary fusion between multi-layer features; and (3) SCF, used to adaptively filter key contextual information. The overall process is shown in Fig. \ref{fig2}, and each component will be described in detail in subsequent sections. Each image $x_i$ is first processed by a Transformer-based encoder to extract multi-level spatial features $F_t\in R^{H\times W\times256},t\in\{1,2,3,4\}$. 
\subsection{Multi-Frequency Decoupling}
Transformer-based architectures have demonstrated significant advantages in long-range dependency modeling for medical image segmentation due to their global receptive field and self-attention mechanism. However, their feature representation remains primarily limited to the spatial domain. In one-shot learning scenarios, limited training data easily leads to model overfitting, and purely spatial domain features cannot adequately decouple global periodic structures, multi-scale local textures, and multi-directional orientation information in semantics, thus limiting the diversity and generalization ability of feature representation. To address these limitations, we propose a Multi-Frequency Decoupling (MFD) module, which simultaneously maps features to the Fourier, wavelet, and Gabor domains.
\subsubsection{Fourier Domain}
The Fourier domain focuses on modeling  periodic patterns and global structure. By explicitly encoding low-frequency semantic contours and long-range dependencies in the frequency space, this design compensates for the spatial domain's limitations in modeling global structure, thereby enhancing the model's ability to stably perceive overall morphology. First, the input feature map $F_4$ is transformed to the frequency domain using a Fast Fourier Transform (FFT) and decomposed into amplitude spectra $\mathrm{\Phi}$ and phase spectra $\theta$. Then, a set of learnable frequency weights $W$ is introduced, constrained by a hyperbolic tangent function, and shifted to the (0,2] interval to perform channel-level modulation of the amplitude spectrum. The modulated amplitude spectrum is recombinated with the original phase spectrum, and the spatial domain features are recovered using an inverse Fourier transform. Finally, a $1\times1$ convolution is performed for channel projection to obtain the output of the Fourier branch. The above formula is:
\begin{equation}F_4^{F}=Con\nu_{1\times1}(f_{IFFT}(\Phi\cdot(1+\tanh(W))\cdot e^{j\theta}))\end{equation}
where $\Phi=\left|f_{FFT}(F_4)\right|$, $\theta=\angle f_{FFT}(F_4)$, $f_{FFT}$ and $f_{IFFT}$ represent the Fast Fourier Transform and its inverse transform, respectively, and $\cdotp$ represents element-wise multiplication.
\subsubsection{Wavelet Domain}
The wavelet domain focuses on texture mining and multi-scale local feature. By decoupling features into different frequency sub-bands, the model can independently model edge, texture, and detail variations at different scales, effectively mitigating the spatial domain's feature aliasing problem. First, perform Haar wavelet transform on the input feature map $F_4$ to decompose the spatial features into four frequency sub-bands: low-frequency approximation component $F_{LL}$, horizontal high-frequency component $F_{LH}$, vertical high-frequency component $F_{HL}$, and diagonal high-frequency component $F_{HH}$. Second, each subband is sequentially processed through depthwise separable convolutions for spatial feature extraction and channel transformation:
\begin{equation}\hat{F}_s=Conv_{1\times1}(\mathrm{DWConv}_{3\times3}(F_s))\end{equation}
where $s\in\{LL,LH,HL,HH\}$. Then, the four encoded subbands are concatenated along the channel dimension to obtain $\hat{F}_s^c$. To adaptively adjust the contribution of different frequency bands to the final result, a subband attention mechanism is introduced: first, spatial information is aggregated through global average pooling, and then the weight coefficients $\alpha_{s}$ of the four subbands are generated through two layers of $1\times1$ convolution and Softmax normalization:
\begin{equation}\alpha_s=S(Con\nu_{1\times1}(ReLU(Con\nu_{1\times1}(GAP(\hat{F}_s^c)))))\end{equation}
where S represents Softmax. Finally, after weighting each subband, they are reconcatenated along the channel dimension, and the weighted multi-frequency band features are fused through $3\times3$ convolution:
\begin{equation}F_4^W=Con\nu_{3\times3}(Concat(a_s\cdot\hat{F}_s))\end{equation}
\subsubsection{Gabor Domain}
The Gabor domain focuses on modeling multi-directional structural response information. Compared to standard convolution relying on data-driven implicit learning of directional information, this design, by injecting directional priors, exhibits stronger stability under low-sample conditions. First, a set of Gabor filter cores $k_G$ with different orientations are constructed. Specifically, a $7\times 7$ coordinate grid is generated on a spatial grid, with the kernel center as the origin, and a total of $N=8$ different directions of $k_G$ are generated, with each kernel corresponding to an angle:
\begin{equation}\lambda_i=\frac{i\cdot\pi}{N},i=0,1,...,N-1\end{equation}

Next, each $k_G$ is normalized to eliminate amplitude differences between different kernels. Then, $k_G$ in different directions are initialized as convolution kernels in the convolutional layer weights. Specifically, for the output channel, its corresponding convolution kernel is filled with a $k_G$ in a fixed direction; at the same time, to ensure consistency between different input channels, the $k_G$ in this direction is shared and copied across all input channels. This design allows each output channel to focus on extracting texture responses in a specific direction.

Finally, after the input feature map $F_4$ undergoes a convolution operation with the Gabor filter kernel, the output multi-directional response feature map $F_4^G$ is obtained:
\begin{equation}F_4^G=F_{4,j}*G_{\lambda_i}\end{equation}
where $*$ represents the convolution operation and $G_{\lambda_i}$ represents the Gabor kernel with a direction angle of $\lambda_i$, and $j$ represents the index of the input channel.\\
\indent We element-wise add the refined features from the three sources to achieve the fusion of complementary information from multiple domains. The fused features are then channel-projected through a $1\times 1$ convolutional layer. Finally, the projected multi-domain fused feature $F_4^{F,W,G}$ is residually connected to the original feature $F_4$ to obtain the final enhanced multi-information feature $F_{4}^{\prime}$:
\begin{equation}F_4^{\prime\prime}=Con\nu_{1\times1}(A(F_4^F)+A(F_4^W)+A(F_4^G))+A(F_4)\end{equation}
where A represents spatial and channel attention mechanisms.
\subsection{Progressive Feature Integrator}
In multi-level feature fusion, features from different layers often exhibit significant distribution differences and expression biases. Shallow features emphasize local texture details, while deep features emphasize global semantic information. Existing fusion strategies typically lack explicit interaction modeling mechanisms, which can easily lead to one level of feature dominating the fusion process and inhibiting the effective expression of information from other levels. To address this limitation, we propose a Progressive Feature Integrator (PFI) module. By constructing a bidirectional interaction mechanism between cross-level features, it guides features from different levels to adaptively select information and complement each other, thereby achieving progressive fusion and balanced expression. Specifically, for the two input feature maps F1 and F2, we first perform average pooling and max pooling to extract the global response for each channel. Second, we perform a non-linear mapping to obtain the channel attention descriptors:
\begin{equation}A_1=MLP(P_{a\nu g}(F_1))+MLP(P_{\max}(F_1))
\end{equation}
\begin{equation}
A_2=MLP(P_{a\nu g}(F_2))+MLP(P_{\max}(F_2))\end{equation}
where $P_{a\nu g}$ and $P_{\max}$ represent average pooling and max pooling, respectively.

Then, we construct a cross-feature correlation matrix $S$ using channel descriptors and generate bidirectional attention weights $A_{1\to2}$ and $A_{2\to1}$ respectively. Finally, we apply the attention weights to the original features and perform residual connections to preserve the original information:
\begin{equation}\hat{F}_1=F_1\cdot Softmax(A_1\cdot A_2^T)+F_1\end{equation}
\begin{equation}\hat{F}_2=F_2\cdot Softmax(A_1\cdot A_2^T)^T+F_2\end{equation}
where T represents transpose. Finally, the two enhanced features are concatenated along the channel dimension and fused using a $1\times 1$ convolution to obtain the final joint feature $F_{1}^{\prime}$:
\begin{equation}F_1^{^{\prime}}=Con\nu_{1\times1}(Concat(\hat{F}_1,\hat{F}_2))\end{equation}

The operations for inputting $F_{3}$ and $F_{4}^{\prime\prime}$ into the PFI are the same as those for inputting $F_{1}$ and $F_{2}$.
\subsection{Selective Context Filtering}
In medical imaging, lesion areas typically occupy only a small proportion of the image, while background areas often contain a large amount of structurally complex and semantically redundant information. Traditional feature extraction and fusion methods usually model within a unified feature space, lacking explicit distinction between foreground and background. This can easily lead to background noise interfering with foreground discrimination, especially in one-shot scenarios where this problem is further amplified. To address this, we propose the Selective Context Filtering (SCF) module. By explicitly decomposing the input features into foreground and background components and constructing independent enhancement branches for each, the network is guided to adaptively strengthen key semantic regions during feature learning while suppressing irrelevant background interference, thereby achieving more accurate and robust lesion localization. Specifically, for the input feature map $F_{4}^{\prime\prime}$, it is first subjected to a $1\times 1$ convolution and then normalized using the Sigmoid function to generate a foreground probability map. Then, the feature map $F_{4}^{\prime\prime}$ is weighted to obtain the enhanced high-confidence foreground region $F_{4}^{\prime\prime\prime}$, and the background region $\hat{F}_4^{\prime\prime\prime\prime}$ is constructed through the complementary relationship of $F_{4}^{\prime\prime\prime}$:
\begin{equation}\hat{F}_4^{\prime\prime\prime\prime}=1-sigmoid(F_4^{\prime\prime\prime})\end{equation}

Next, the foreground branch $F_4^{fg}$ enhances the semantic representation of the foreground region through continuous convolutional operations, while the background branch $F_4^{bg}$ uses the same network structure to suppress the background region:
\begin{equation}F_4^{fg}=\Pi(sigmoid(F_4^{\prime\prime\prime})),F_4^{bg}=\Pi(\hat{F}_4^{\prime\prime\prime\prime})\end{equation}
where $\Pi()$ represents a continuous convolution operation. 

Finally, attention mechanisms are applied to $F_4^{fg}$ and $F_4^{bg}$ respectively, and residual connections are made with $F_{4}^{\prime\prime\prime\prime}$ and $\hat{F}_4^{\prime\prime\prime\prime}$. This allows the foreground to focus on the key responses of the target region, while the background strengthens the identification and suppression of interference patterns, thereby achieving differentiated optimization at the feature level. Then, $F_4^{fg}$ and $F_4^{bg}$ are merged to generate the output $F_{4}^{\prime\prime\prime\prime\prime\prime}$.

The result obtained by fusing $F_{1}^{\prime}$ and $F_{3}^{\prime}$ and inputting it into the SCF module is then fused with $F_{4}^{\prime\prime\prime\prime\prime\prime}$ to form the final output $F_{1}^{\prime\prime}$ of the SAM image encoder.
\subsection{Loss Function and Evaluation Metrics}
The loss function uses a weighted sum of Binary Cross-Entropy (BCE) loss and Dice loss to achieve optimal segmentation performance:
\begin{equation}\mathcal{L}_{OB-SAM}=\mathcal{L}_{BCE}+\mathcal{L}_{Dice},\end{equation}

We use Dice \cite{yu2019uncertainty} and Intersection-over-Union (IoU) \cite{wu2022mutual} to measure pixel-level consistency between the predicted results and the actual values.

\begin{figure*}[!t]
\centerline{\includegraphics[width=2.10\columnwidth]{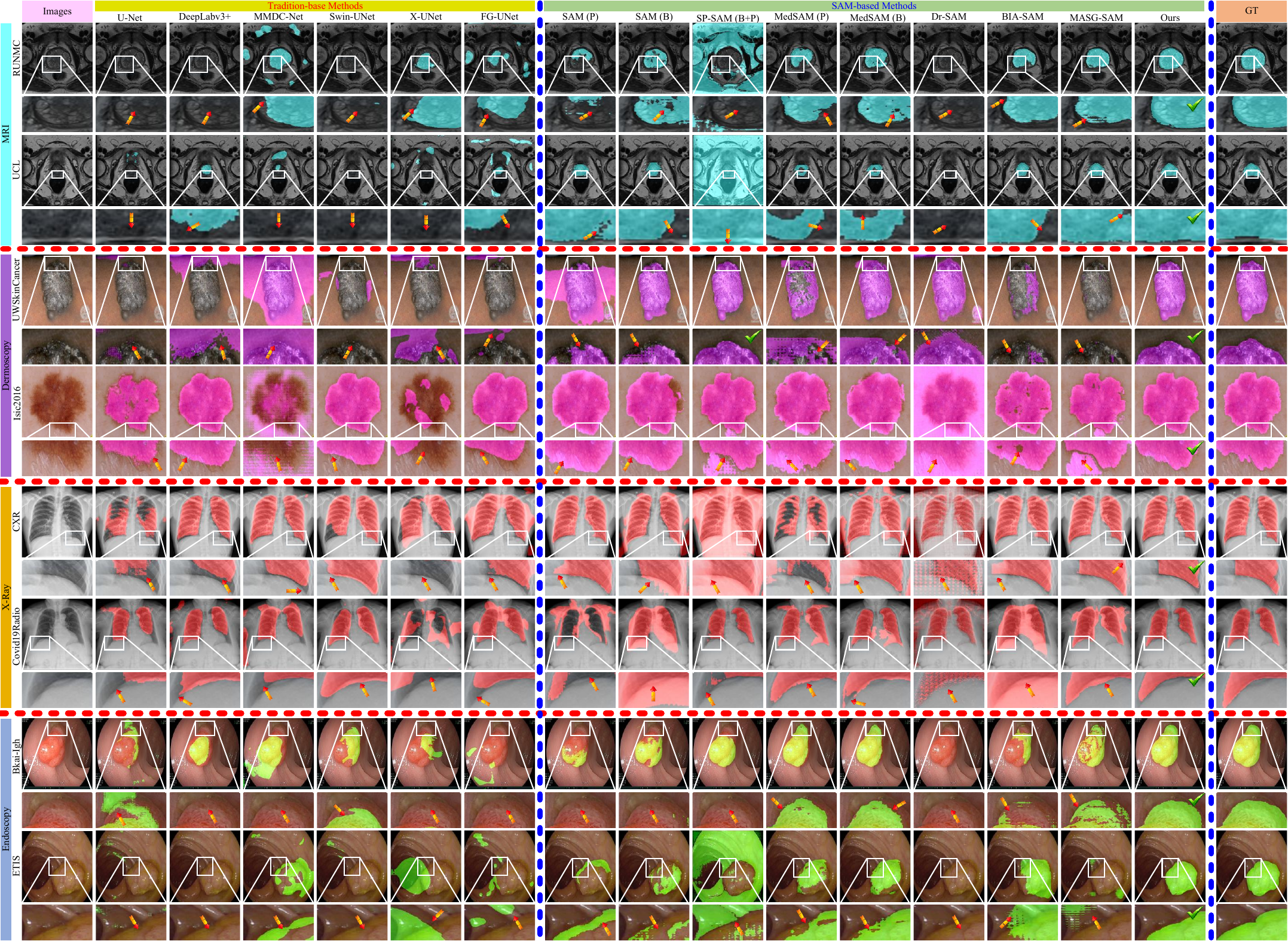}}
\caption{Qualitative visualization comparison of segmentation results from different methods on seven medical datasets is provided.}
\label{fig4}
\end{figure*}
\section{Experiments}
\subsection{Datasets}
\begin{table}[]
\caption{Details of the datasets for experiments.}
\scalebox{0.58}{
\begin{tabular}{ccccc}
\hline
\Xhline{1.1pt}
\rowcolor[HTML]{EAEAEA} 
Datasets Name & Modality                      & Train Quantity & Test Quantity & \begin{tabular}[c]{@{}c@{}}OB-SAM use quantity/\\ proportion(\%)\end{tabular} \\ \hline
RUNMC \cite{liu2020shape}         & MRI                           & 338            & 83            & 1/0.296                                                                      \\
UCL \cite{liu2020shape}           & MRI                           & 133            & 42            & 1/0.752                                                                      \\
UWSkinCancer \cite{glaister2013automatic}  & Dermoscopy                    & 143            & 44            & 1/0.699                                                                      \\
ISIC2016 \cite{codella2018skin}      & Dermoscopy                    & 900            & 379           & 1/0.111                                                                      \\
CXR \cite{candemir2013lung}          & Chest X-ray                   & 492            & 212           & 1/0.203                                                                      \\
Covid19Radio \cite{rahman2021exploring}  & Chest X-ray                   & 14814          & 4236          & 1/0.006                                                                      \\
Bkai-Igh \cite{ngoc2021neounet}      & \multicolumn{1}{l}{Endoscopy} & 700           & 200           & 1/0.100                                                                      \\
ETIS \cite{silva2014toward}         & Endoscopy                     & 137            & 59            & 1/0.730                                                                      \\ \hline
\Xhline{1.1pt}
\end{tabular}
}
\label{tab1}
\end{table}
This paper uses seven publicly available datasets covering four imaging modalities: MRI: two subsets, UCL and RUNMC, from the prostate dataset \cite{liu2020shape}, with the partitioning following \cite{ma2025steady}. Dermoscopy: UWSkinCancer \cite{glaister2013automatic} and ISIC2016 \cite{codella2018skin}, the former partitioned according to \cite{kucs2024medsegbench}, and the latter using the official train/test partitioning. Chest X-rays: CXR \cite{candemir2013lung} (train/test 7:3) and Covid19Radio \cite{rahman2021exploring} (partitioning according to \cite{kucs2024medsegbench}). Endoscopy: Bkai-Igh \cite{ngoc2021neounet} (partitioning method same as \cite{kucs2024medsegbench}) and ETIS \cite{silva2014toward} (train/test 7:3). The names, modalities, train/test set sample sizes, and experimental scales of each dataset are detailed in Table \ref{tab1}. 

\subsection{Implementation Details}

Our network leverages the Transformer architecture, initialized with the pre-trained ViT-B model from SAM, and trained for 100 epochs with a batch size of 2. All experiments are conducted on an NVIDIA GeForce RTX 4090 GPU. The optimizer uses AdamW with parameters set to $\beta_{1}$ = 0.9, $\beta_{2}$ = 0.999, an initial learning rate of 5e-3, and a weight decay coefficient of 0.01. For image preprocessing, all images and corresponding labels are resized to a resolution of 1024×1024.
\begin{table*}[htbp]
\definecolor{OBSAMScoreBg}{HTML}{FCECEC}
\definecolor{OBSAMSecondBg}{HTML}{E8F1FD}
\caption{One-shot OB-SAM and fully supervised CNN-based methods are compared on seven datasets in terms of Dice (\%) and IoU (\%). \ding{55} = No prompt. OB-SAM scores are shaded in light pink. Best and second-best are shown in \colorbox{OBSAMScoreBg}{Light pink} and \colorbox{OBSAMSecondBg}{light blue}}
\newcommand{\OBSAMHighlight}[2]{%
  \begingroup
  \setlength{\fboxsep}{0pt}%
  \colorbox{#1}{\strut\kern.8pt#2\kern.8pt}%
  \endgroup
}
\newcommand{\OBSAMSecond}[1]{\OBSAMHighlight{OBSAMSecondBg}{#1}}
\newcommand{\OBSAMScore}[1]{\OBSAMHighlight{OBSAMScoreBg}{#1}}
\scalebox{0.605}{
\begin{tabular}{ccc|cccc|cccc|cccc|cccc}
\Xhline{1.1pt}
\hline
\rowcolor[HTML]{EAEAEA} 
\cellcolor[HTML]{EAEAEA}                         & \cellcolor[HTML]{EAEAEA}                        & \cellcolor[HTML]{EAEAEA}                         & \multicolumn{4}{c|}{\cellcolor[HTML]{EAEAEA}MRI}                                                     & \multicolumn{4}{c|}{\cellcolor[HTML]{EAEAEA}Dermoscopy}                                                            & \multicolumn{4}{c|}{\cellcolor[HTML]{EAEAEA}Chest X-ray}                                                    & \multicolumn{4}{c}{\cellcolor[HTML]{EAEAEA}Endoscopy}                                                   \\ \cline{4-19} 
\rowcolor[HTML]{EAEAEA} 
\cellcolor[HTML]{EAEAEA}                         & \cellcolor[HTML]{EAEAEA}                        & \cellcolor[HTML]{EAEAEA}                         & \multicolumn{2}{c}{\cellcolor[HTML]{EAEAEA}RUNMC \cite{liu2020shape}} & \multicolumn{2}{c|}{\cellcolor[HTML]{EAEAEA}UCL \cite{liu2020shape}} & \multicolumn{2}{c}{\cellcolor[HTML]{EAEAEA}UWSkinCancer \cite{glaister2013automatic}} & \multicolumn{2}{c|}{\cellcolor[HTML]{EAEAEA}ISIC2016 \cite{codella2018skin}} & \multicolumn{2}{c}{\cellcolor[HTML]{EAEAEA}CXR \cite{candemir2013lung}} & \multicolumn{2}{c|}{\cellcolor[HTML]{EAEAEA}Covid19Radio \cite{rahman2021exploring}} & \multicolumn{2}{c}{\cellcolor[HTML]{EAEAEA}Bkai-Igh \cite{ngoc2021neounet}} & \multicolumn{2}{c}{\cellcolor[HTML]{EAEAEA}ETIS \cite{silva2014toward}} \\ \cline{4-19} 
\rowcolor[HTML]{EAEAEA} 
\multirow{-3}{*}{\cellcolor[HTML]{EAEAEA}Method} & \multirow{-3}{*}{\cellcolor[HTML]{EAEAEA}Venue} & \multirow{-3}{*}{\cellcolor[HTML]{EAEAEA}Prompt} & Dice                    & IoU                     & Dice                    & IoU                    & Dice                         & IoU                         & Dice                      & IoU                       & Dice                   & IoU                    & Dice                        & IoU                         & Dice                      & IoU                      & Dice                    & Iou                    \\ \hline
U-Net \cite{ronneberger2015u}                                            & MICCAI’15                                       & \ding{55}                                                & 75.63                   & 65.51                   & 28.15                   & 20.59                  & 87.92                        & 80.98                       & 90.60                     & 83.96                     & \textbf{95.31}         & \textbf{91.15}         & 97.34                       & 94.95                       & 72.32                     & 64.07                    & 28.51                   & 19.84                  \\
DeepLabv3+ \cite{chen2018encoder}                                       & ECCV’18                                         & \ding{55}                                                & \OBSAMSecond{82.80}             & \OBSAMSecond{73.64}             & \OBSAMSecond{82.80}             & \OBSAMSecond{73.64}            & \OBSAMSecond{91.22}                  & \OBSAMSecond{85.04}                 & \textbf{90.87}            & \textbf{84.22}            & 94.07                  & 89.04                  & 97.26                       & 94.81                       & 82.59                     & 76.81                    & \OBSAMSecond{77.66}             & \OBSAMSecond{69.10}            \\
MMDC-Net \cite{zhong2022you}                                         & CIBM’22                                         & \ding{55}                                                & 65.65                   & 55.88                   & 72.98                   & 62.60                  & 71.13                        & 60.88                       & 87.19                     & 79.32                     & 94.83                  & 90.38                  & \textbf{98.10}              & \textbf{96.38}              & 69.69                     & 61.18                    & 39.55                   & 29.49                  \\
Swin-UNet \cite{cao2022swin}                                        & ECCV’22                                         & \ding{55}                                                & 79.70                   & 69.58                   & 0.78                    & 0.41                   & 89.34                        & 82.80                       & \OBSAMSecond{90.72}               & \OBSAMSecond{84.20}               & 95.09                  & 90.82                  & 96.18                       & 92.88                       & 75.02                     & 66.44                    & 0.66                    & 0.35                   \\
X-UNet \cite{xu2025x}                                           & NN’25                                           & \ding{55}                                                & 79.94                   & 71.50                   & 70.11                   & 59.86                  & 87.48                        & 80.50                       & 89.70                     & 83.14                     & \OBSAMSecond{95.25}            & \OBSAMSecond{91.07}            & 97.80                       & 95.87                       & \OBSAMSecond{84.81}               & \OBSAMSecond{78.26}              & 53.68                   & 43.33                  \\
FG-UNet \cite{wang2025rethinking}                                          & IEEE SPL’25                                     & \ding{55}                                                & 78.81                   & 69.37                   & 79.98                   & 69.71                  & 88.30                        & 80.89                       & 89.93                     & 83.29                     & 95.19                  & 90.95                  & \OBSAMSecond{97.88}                 & \OBSAMSecond{96.01}                 & 77.43                     & 70.36                    & 1.08                    & 0.60                   \\ \hline
\textbf{Ours (OB-SAM)}                                              & \textbf{This paper}                                      & \ding{55}                                                & \OBSAMScore{86.88}          & \OBSAMScore{77.96}          & \OBSAMScore{85.37}          & \OBSAMScore{75.81}         & \OBSAMScore{91.88}               & \OBSAMScore{85.38}              & \OBSAMScore{85.83}                     & \OBSAMScore{76.57}                     & \OBSAMScore{90.98}                  & \OBSAMScore{83.61}                  & \OBSAMScore{84.48}                       & \OBSAMScore{74.28}                       & \OBSAMScore{86.90}            & \OBSAMScore{79.67}           & \OBSAMScore{81.42}          & \OBSAMScore{71.18}         \\ \hline
\Xhline{1.1pt}
\end{tabular}
}
\label{tab2}
\end{table*}

\begin{table*}[htbp]
\definecolor{OBSAMScoreBg}{HTML}{FCECEC}
\definecolor{OBSAMSecondBg}{HTML}{E8F1FD}
\caption{Comparison of One-shot
OB-SAM and CNN-based methods. "-" indicates poor performance on this dataset.}
\newcommand{\OBSAMHighlight}[2]{%
  \begingroup
  \setlength{\fboxsep}{0pt}%
  \colorbox{#1}{\strut\kern.8pt#2\kern.8pt}%
  \endgroup
}
\newcommand{\OBSAMSecond}[1]{\OBSAMHighlight{OBSAMSecondBg}{#1}}
\newcommand{\OBSAMScore}[1]{\OBSAMHighlight{OBSAMScoreBg}{#1}}
\scalebox{0.605}{
\begin{tabular}{ccc|cccc|cccc|cccc|cccc}
\Xhline{1.1pt}
\hline
\rowcolor[HTML]{EAEAEA} 
\cellcolor[HTML]{EAEAEA}                         & \cellcolor[HTML]{EAEAEA}                        & \cellcolor[HTML]{EAEAEA}                         & \multicolumn{4}{c|}{\cellcolor[HTML]{EAEAEA}MRI}                                                     & \multicolumn{4}{c|}{\cellcolor[HTML]{EAEAEA}Dermoscopy}                                                            & \multicolumn{4}{c|}{\cellcolor[HTML]{EAEAEA}Chest X-ray}                                                    & \multicolumn{4}{c}{\cellcolor[HTML]{EAEAEA}Endoscopy}                                                   \\ \cline{4-19} 
\rowcolor[HTML]{EAEAEA} 
\cellcolor[HTML]{EAEAEA}                         & \cellcolor[HTML]{EAEAEA}                        & \cellcolor[HTML]{EAEAEA}                         & \multicolumn{2}{c}{\cellcolor[HTML]{EAEAEA}RUNMC \cite{liu2020shape}} & \multicolumn{2}{c|}{\cellcolor[HTML]{EAEAEA}UCL \cite{liu2020shape}} & \multicolumn{2}{c}{\cellcolor[HTML]{EAEAEA}UWSkinCancer \cite{glaister2013automatic}} & \multicolumn{2}{c|}{\cellcolor[HTML]{EAEAEA}ISIC2016 \cite{codella2018skin}} & \multicolumn{2}{c}{\cellcolor[HTML]{EAEAEA}CXR \cite{candemir2013lung}} & \multicolumn{2}{c|}{\cellcolor[HTML]{EAEAEA}Covid19Radio \cite{rahman2021exploring}} & \multicolumn{2}{c}{\cellcolor[HTML]{EAEAEA}Bkai-Igh \cite{ngoc2021neounet}} & \multicolumn{2}{c}{\cellcolor[HTML]{EAEAEA}ETIS \cite{silva2014toward}} \\ \cline{4-19} 
\rowcolor[HTML]{EAEAEA} 
\multirow{-3}{*}{\cellcolor[HTML]{EAEAEA}Method} & \multirow{-3}{*}{\cellcolor[HTML]{EAEAEA}Venue} & \multirow{-3}{*}{\cellcolor[HTML]{EAEAEA}Prompt} & Dice                    & IoU                     & Dice                    & IoU                    & Dice                         & IoU                         & Dice                      & IoU                       & Dice                   & IoU                    & Dice                        & IoU                         & Dice                      & IoU                      & Dice                    & Iou                    \\ \hline
U-Net \cite{ronneberger2015u}                                            & MICCAI’15                                       & \ding{55}                                                & -                       & -                       & 5.19                    & 2.88                   & 55.41                        & 44.38                       & 68.62                     & 57.26                     & 64.10                  & 49.29                  & 57.93                       & 44.96                       & 14.15                     & 8.67                     & 3.56                    & 2.35                   \\
DeepLabv3+ \cite{chen2018encoder}                                       & ECCV’18                                         & \ding{55}                                                & -                       & -                       & 5.47                    & 3.28                   & \OBSAMSecond{58.47}                  & \OBSAMSecond{48.91}                 & \OBSAMSecond{71.91}               & \OBSAMSecond{60.92}               & \OBSAMSecond{88.64}            & \OBSAMSecond{80.21}            & \OBSAMSecond{81.27}                 & \OBSAMSecond{69.61}                 & 16.04                     & 10.92                    & 1.47                    & 0.81                   \\
MMDC-Net \cite{zhong2022you}                                         & CIBM’22                                         & \ding{55}                                                & 17.58                   & 11.17                   & 1.70                    & 0.92                   & 44.66                        & 34.32                       & 30.60                     & 21.22                     & 80.90                  & 69.95                  & 65.87                       & 52.66                       & \OBSAMSecond{19.51}               & \OBSAMSecond{12.23}              & \OBSAMSecond{9.59}              & \OBSAMSecond{6.31}             \\
Swin-UNet \cite{cao2022swin}                                        & ECCV’22                                         & \ding{55}                                                & 5.40                    & 3.09                    & 1.30                    & 0.68                   & 53.91                        & 41.26                       & 66.86                     & 56.19                     & 82.90                  & 72.18                  & 79.55                       & 67.28                       & 16.33                     & 10.85                    & 0.03                    & 0.01                   \\
X-UNet \cite{xu2025x}                                           & NN’25                                           & \ding{55}                                                & \OBSAMSecond{19.20}             & \OBSAMSecond{12.18}             & \OBSAMSecond{18.30}             & \OBSAMSecond{11.98}            & 34.64                        & 24.87                       & 45.30                     & 33.05                     & 71.12                  & 56.91                  & 57.47                       & 41.93                       & 13.78                     & 8.86                     & 8.65                    & 5.66                   \\
FG-UNet \cite{wang2025rethinking}                                          & IEEE SPL’25                                     & \ding{55}                                                & 14.01                   & 8.44                    & 3.46                    & 1.84                   & 41.91                        & 29.86                       & 59.29                     & 47.21                     & 69.30                  & 54.14                  & 79.40                       & 66.94                       & 8.08                      & 4.62                     & 3.51                    & 1.96                   \\ \hline
\textbf{Ours (OB-SAM)}                                              & \textbf{This paper}                                      & \ding{55}                                                & \OBSAMScore{86.88}          & \OBSAMScore{77.96}          & \OBSAMScore{85.37}          & \OBSAMScore{75.81}         & \OBSAMScore{91.88}               & \OBSAMScore{85.38}              & \OBSAMScore{85.38}            & \OBSAMScore{76.57}            & \OBSAMScore{90.98}         & \OBSAMScore{83.61}         & \OBSAMScore{84.48}              & \OBSAMScore{74.28}              & \OBSAMScore{86.90}            & \OBSAMScore{79.67}           & \OBSAMScore{81.42}          & \OBSAMScore{71.18}         \\ \hline
\Xhline{1.1pt}
\end{tabular}
}
\label{tab3}
\end{table*}

\begin{table*}[!t]
\definecolor{OBSAMScoreBg}{HTML}{FCECEC}
\definecolor{OBSAMSecondBg}{HTML}{E8F1FD}
\caption{Comparison of One-shot OB-SAM and SAM-base methods. P = Point prompt, B = Box prompt, B+P = combined prompt.}
\newcommand{\OBSAMHighlight}[2]{%
  \begingroup
  \setlength{\fboxsep}{0pt}%
  \colorbox{#1}{\strut\kern.8pt#2\kern.8pt}%
  \endgroup
}
\newcommand{\OBSAMSecond}[1]{\OBSAMHighlight{OBSAMSecondBg}{#1}}
\newcommand{\OBSAMScore}[1]{\OBSAMHighlight{OBSAMScoreBg}{#1}}
\scalebox{0.595}{
\begin{tabular}{ccc|cccc|cccc|cccc|cccc}
\Xhline{1.1pt}
\hline
\rowcolor[HTML]{EAEAEA} 
\cellcolor[HTML]{EAEAEA}                         & \cellcolor[HTML]{EAEAEA}                        & \cellcolor[HTML]{EAEAEA}                         & \multicolumn{4}{c|}{\cellcolor[HTML]{EAEAEA}MRI}                                                     & \multicolumn{4}{c|}{\cellcolor[HTML]{EAEAEA}Dermoscopy}                                                            & \multicolumn{4}{c|}{\cellcolor[HTML]{EAEAEA}Chest X-ray}                                                    & \multicolumn{4}{c}{\cellcolor[HTML]{EAEAEA}Endoscopy}                                                   \\ \cline{4-19} 
\rowcolor[HTML]{EAEAEA} 
\cellcolor[HTML]{EAEAEA}                         & \cellcolor[HTML]{EAEAEA}                        & \cellcolor[HTML]{EAEAEA}                         & \multicolumn{2}{c}{\cellcolor[HTML]{EAEAEA}RUNMC \cite{liu2020shape}} & \multicolumn{2}{c|}{\cellcolor[HTML]{EAEAEA}UCL \cite{liu2020shape}} & \multicolumn{2}{c}{\cellcolor[HTML]{EAEAEA}UWSkinCancer \cite{glaister2013automatic}} & \multicolumn{2}{c|}{\cellcolor[HTML]{EAEAEA}ISIC2016 \cite{codella2018skin}} & \multicolumn{2}{c}{\cellcolor[HTML]{EAEAEA}CXR \cite{candemir2013lung}} & \multicolumn{2}{c|}{\cellcolor[HTML]{EAEAEA}Covid19Radio \cite{rahman2021exploring}} & \multicolumn{2}{c}{\cellcolor[HTML]{EAEAEA}Bkai-Igh \cite{ngoc2021neounet}} & \multicolumn{2}{c}{\cellcolor[HTML]{EAEAEA}ETIS \cite{silva2014toward}} \\ \cline{4-19} 
\rowcolor[HTML]{EAEAEA} 
\multirow{-3}{*}{\cellcolor[HTML]{EAEAEA}Method} & \multirow{-3}{*}{\cellcolor[HTML]{EAEAEA}Venue} & \multirow{-3}{*}{\cellcolor[HTML]{EAEAEA}Prompt} & Dice                    & IoU                     & Dice                    & IoU                    & Dice                         & IoU                         & Dice                      & IoU                       & Dice                   & IoU                    & Dice                        & IoU                         & Dice                      & IoU                      & Dice                    & Iou                    \\ \hline
SAM \cite{kirillov2023segment}                                              & ICCV’23                                         & P                                                & 67.54                   & 54.96                   & \OBSAMSecond{82.89}             & \OBSAMSecond{71.98}            & 21.93                        & 13.46                       & 77.28                     & 66.28                     & \OBSAMSecond{90.80}            & \OBSAMScore{83.65}         & 47.50                       & 32.67                       & 66.12                     & 54.59                    & 61.85                   & 51.12                  \\
SAM \cite{kirillov2023segment}                                            & ICCV’23                                         & B                                                & 68.81                   & 57.17                   & 78.17                   & 66.37                  & 87.12                        & 77.98                       & 83.08                     & 73.19                     & 71.67                  & 56.75                  & 64.97                       & 48.92                       & 79.45                     & 69.59                    & 63.41                   & 51.81                  \\
SP-SAM \cite{wu2023self}                                           & MICCAI’23                                       & P+B                                              & 0.31                    & 0.16                    & 51.67                   & 41.89                  & \OBSAMSecond{89.10}                  & \OBSAMSecond{82.02}                & 64.76                     & 54.60                     & 55.00                  & 40.30                  & 67.47                       & 54.98                       & 49.53                     & 44.63                    & 19.51                   & 16.04                  \\
MedSAM \cite{ma2024segment}                                           & Nat.Commun’24                                   & P                                                & 71.68                   & 57.51                   & 67.97                   & 52.79                  & 83.43                        & 72.76                       & \OBSAMSecond{84.04}               & \OBSAMSecond{73.78}               & 81.14                  & 69.54                  & \OBSAMSecond{84.28}                 & \OBSAMSecond{73.41}                 & 82.44                     & 72.07                    & 76.71                   & 64.23                  \\
MedSAM \cite{ma2024segment}                                           & Nat.Commun’24                                   & B                                                & 78.10                   & 65.53                   & 76.99                   & 63.47                  & 86.82                        & 77.31                       & 78.78                     & 65.87                     & 73.62                  & 58.58                  & 81.66                       & 69.89                       & \OBSAMSecond{86.31}               & \OBSAMSecond{77.74}              & \OBSAMSecond{78.28}             & \OBSAMSecond{66.43}            \\
Dr-SAM \cite{huo2024dr}                                           & ICIC’24                                         & \ding{55}                                                & -                       & -                       & -                       & -                      & 66.36                        & 53.62                       & 41.59                     & 29.70                     & 49.05                  & 32.87                  & 59.72                       & 43.31                       & -                         & -                        & -                       & -                      \\
BIA-SAM \cite{zhou2024biasam}                                          & IEEE SPL’25                                     & P+B                                              & 81.13                   & 69.30                   & 81.22                   & 70.42                  & 67.89                        & 57.79                       & 75.45                     & 63.24                     & 86.61                  & 76.83                  & 65.59                       & 49.70                       & 67.27                     & 55.17                    & 71.98                   & 58.58                  \\
MASG-SAM \cite{zhou2025masg}                                         & IEEE JBHI’25                                    & B                                                & \OBSAMSecond{84.00}             & \OBSAMSecond{74.42}             & 81.88                   & 70.82                  & 58.71                        & 46.01                       & 78.87                     & 68.28                     & 90.27                  & 82.55                  & 82.93                       & 71.39                       & 81.90                     & 71.64                    & 45.49                   & 33.38                  \\ \hline
\textbf{Ours (OB-SAM)}                           & \textbf{This paper}                             & B                                                & \OBSAMScore{86.88}          & \OBSAMScore{77.96}          & \OBSAMScore{85.37}          & \OBSAMScore{75.81}         & \OBSAMScore{91.88}               & \OBSAMScore{85.38}              & \OBSAMScore{85.83}            & \OBSAMScore{76.57}            & \OBSAMScore{90.98}         & \OBSAMSecond{83.61}            & \OBSAMScore{84.48}              & \OBSAMScore{74.28}              & \OBSAMScore{86.90}            & \OBSAMScore{79.67}           & \OBSAMScore{81.42}          & \OBSAMScore{71.18}         \\ \hline
\Xhline{1.1pt}
\end{tabular}
}
\label{tab4}
\end{table*}

\subsection{Comparison to State-of-the-art}

We compare OB-SAM with representative medical image segmentation methods, including classic CNN-based models (U-Net \cite{ronneberger2015u}, DeepLabv3+ \cite{chen2018encoder}, MMDC-Net \cite{zhong2022you}, Swin-UNet \cite{cao2022swin}, X-UNet \cite{xu2025x}, FG-UNet \cite{wang2025rethinking}) and recent SAM-based adaptations (SAM \cite{kirillov2023segment}, SP-SAM \cite{wu2023self}, MedSAM \cite{ma2024segment} , Dr-SAM \cite{huo2024dr}, BIA-SAM \cite{zhou2024biasam}, MASG-SAM \cite{zhou2025masg}). To ensure fair comparison, all methods were evaluated under the same experimental protocol, covering eight tasks and multiple imaging modalities.

Table \ref{tab2} compares the segmentation performance of One-shot OB-SAM with various fully supervised CNN-based methods on different datasets. Overall, despite using only one labeled sample, OB-SAM achieves comparable or even better performance than fully supervised methods on most datasets. Specifically, on the MRI (RUNMC, UCL) and Endoscopy (Bkai-Igh, ETIS) datasets, its Dice/IoU is improved by 4.08\%/4.32\%, 2.57\%/2.17\%, 2.09\%/1.41\%, and 3.76\%/2.08\% respectively compared to the best-performing comparison method. This result indicates that OB-SAM has stronger adaptability and robustness in low-contrast (MRI) and noisy (Endoscopy) scenarios.

Tables \ref{tab3} and \ref{tab4} compare the segmentation performance of One-shot OB-SAM with CNN-based and SAM-based methods on multiple datasets. Overall, OB-SAM achieves the best performance in all scenarios. In particular, on datasets where CNN-based methods are almost ineffective: RUNMC (Dice/IoU: 19.20\%/12.18\%), UCL (18.30\%/11.98\%), Bkai-Igh (19.51\%/12.23\%), and ETIS (9.59\%/6.31\%), OB-SAM still consistently achieves Dice/IoU performance of 81.42\%/71.18\% – 86.90\%/79.67\%, representing a performance improvement of 67 – 77 percentage points. Compared to SAM-based methods, OB-SAM consistently outperforms all datasets, particularly on the RUNMC and ETIS datasets, where Dice/IoU are 2.88\%/3.54\% and 3.14\%/4.75\% higher, respectively. These results fully validate the effectiveness and superior generalization ability of OB-SAM under extremely low annotation conditions.

The visual results in Fig. \ref{fig4} further support the quantitative findings, demonstrating that OB-SAM generates continuous segmentations that closely align with target organ morphology and region responses that are insensitive to imaging artifacts, showing strong agreement with the ground truth. Notably, on challenging modalities such as Endoscopy and MRI, the comparison methods exhibit varying degrees of region discontinuities or false detections, whereas OB-SAM preserves complete anatomical structures. In summary, OB-SAM achieves performance comparable to fully-supervised CNN-based models while outperforming existing SAM-based methods in the one-shot scenario.
\subsection{Ablation Studies}
\subsubsection{Impact of Different Components}
\begin{figure}[]
\centerline{\includegraphics[width=0.98\columnwidth]{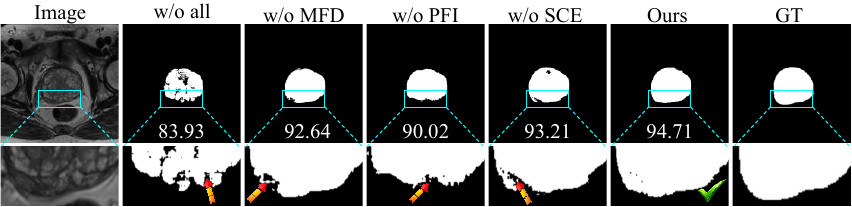}}
\caption{Qualitative ablation results, “w/o” indicates module removal. White shows predictions, blue shows zoomed views.}
\label{fig5}
\end{figure}

To verify the contribution of each core module (MFD, PFI, and SCF) in OB-SAM, we conducted detailed ablation experiments on the RUNMC dataset. Quantitative results are shown in Table \ref{tab5}. First, removing the MFD module resulted in a 3.46\% decrease in Dice coefficient and a 4.89\% decrease in IoU. This significant performance degradation indicates that MFD plays a crucial role in modeling information at different semantic levels. Second, removing the PFI module resulted in a 1.82\% decrease in Dice coefficient and a 2.71\% decrease in IoU. This result demonstrates that PFI promotes progressive fusion of cross-layer features, achieving information complementarity. Finally, removing the SCF module resulted in a 3.06\% decrease in Dice coefficient and a 4.17\% decrease in IoU. This indicates that SCF is significant for capturing key discriminative information, suppressing redundant interference, and achieving accurate localization of lesion regions. Qualitative results (as shown in Fig. \ref{fig5}) further validate the above conclusions. Compared to the complete model, removing any module leads to varying degrees of performance degradation, such as missing local structures. In contrast, the synergistic effect of all three modules yields more accurate and robust segmentation results.
\begin{table}
\caption{Ablation study of different components on RUNMC dataset. \ding{51} indicates the component is enabled.}
\centering
\definecolor{RUNMCBestBg}{HTML}{FCECEC}
\setlength{\fboxsep}{0.8pt}
\scalebox{0.89}{%
  \renewcommand{\arraystretch}{0.90}
  \begin{tabular}{ccc|cc}
    \Xhline{1.1pt}
    \rowcolor[HTML]{EAEAEA}
    MFD Module & PFI Module & SCF Module & Dice & IoU \\
    \hline
    \textbf{-} & \textbf{-} & \textbf{-} & 68.81 & 57.17 \\
    \textbf{-} & \ding{51} & \ding{51} & 83.42 & 73.07 \\
    \ding{51} & \textbf{-} & \ding{51} & 85.06 & 75.25 \\
    \ding{51} & \ding{51} & \textbf{-} & 83.82 & 73.79 \\
    \ding{51} & \ding{51} & \ding{51}
      & \colorbox{RUNMCBestBg}{86.88}
      & \colorbox{RUNMCBestBg}{77.96} \\
    \Xhline{1.1pt}
  \end{tabular}%
}
\label{tab5}
\end{table}
\subsubsection{Visualization comparison of heatmap}
\begin{figure}[]
\centerline{\includegraphics[width=0.98\columnwidth]{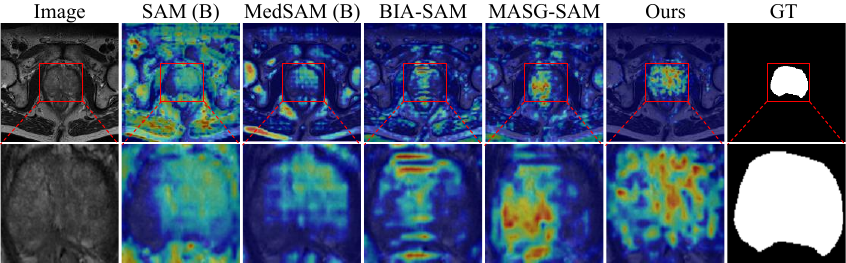}}
\caption{Heatmap visualizations of our method and some state-of-the-art methods.}
\label{fig6}
\end{figure}

To visually evaluate the attention capabilities of different methods to the target region, we generated corresponding heatmaps for each model, as shown in Fig. \ref{fig6}. It can be observed that the original SAM and some contrasting methods often exhibit unstable attention distributions: either the attention region deviates from the target structure, resulting in significant false activations; or the response to the target region is insufficient, failing to fully cover key anatomical structures. In contrast, our method can form a more concentrated and continuous high-response region on the target area, while effectively suppressing background interference. This shows that OB-SAM can obtain more accurate and complete segmentation results.
\begin{figure}[h]
\centerline{\includegraphics[width=0.98\columnwidth]{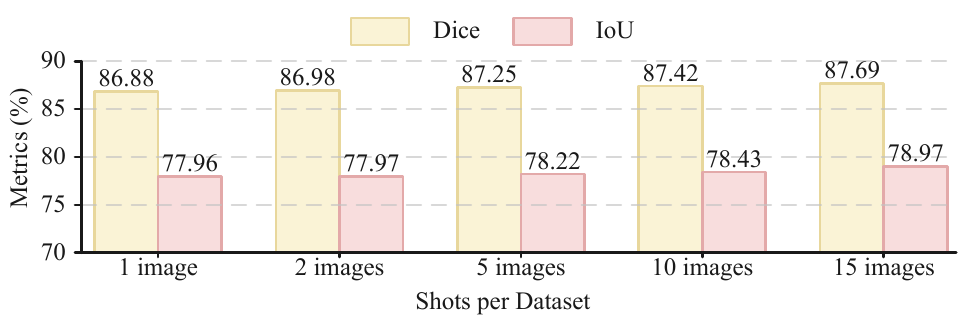}}
\caption{On the RUNMC dataset, the Dice value of OB-SAM changes as the number of training shots increases.}
\label{fig7}
\end{figure}
\subsubsection{Impact of Training Shot Size}
We investigated the data efficiency of OB-SAM by varying the number of training samples in each dataset. Experiments were conducted with settings ranging from 1-shot to 15-shot. As shown in Fig. \ref{fig7}, OB-SAM achieved high performance even with only one training sample, demonstrating its good generalization ability in low-sample scenarios. While the model performance generally increased with the number of samples, the increase was limited, indicating that OB-SAM has learned discriminative feature representations even under extremely low-sample conditions.

\section{Conclusion}
This paper proposes a general medical image segmentation framework, OB-SAM, based on single-sample fine-tuning. OB-SAM effectively mitigates overfitting risk, ensures balanced feature representation, and improves lesion localization capabilities through three structured modules: MFD, PFI, and SCF. Experimental results on seven datasets demonstrate that OB-SAM consistently outperforms state-of-the-art methods with only a single training sample. Future work will further explore lighter and more efficient adaptation strategies, as well as robustness improvements in extreme domain shifts and complex clinical scenarios.

{
    \small
    \bibliographystyle{ieeenat_fullname}
    \bibliography{main}
}

% WARNING: do not forget to delete the supplementary pages from your submission 
% \input{sec/X_suppl}
\end{document}